\documentclass[letterpaper, 10 pt, conference]{ieeeconf}  
\IEEEoverridecommandlockouts                         
\usepackage[T1]{fontenc}
\usepackage{aecompl}
\usepackage{epstopdf}
\usepackage{cite}
\usepackage{amsmath,amssymb,amsfonts}
\usepackage{algorithmic}
\usepackage{algorithm}
\usepackage{graphicx}
\usepackage{textcomp}
\usepackage{xcolor}
\usepackage{float}
\usepackage{graphicx}
\usepackage{picinpar}
\usepackage{babel}
\usepackage{amsmath}
\usepackage{url}
\usepackage[latin1]{inputenc}
\usepackage{colortbl}
\usepackage{soul}
\usepackage{multirow}
\usepackage{pifont}
\usepackage{color}
\usepackage{alltt}
\usepackage[hidelinks]{hyperref}
\usepackage{enumerate}
\usepackage{siunitx}
\usepackage{breakurl}
\usepackage{epstopdf}
\usepackage{pbox}
\usepackage{authblk}
\usepackage{babel}
\usepackage{mathrsfs,balance,subfigure}
\usepackage{tcolorbox}
\usepackage{subfigure}
\usepackage{stfloats}

\title{\LARGE \bf Separated Proportional-Integral Lagrangian for Chance Constrained Reinforcement Learning}

\author{Baiyu Peng$^{1}$, Yao Mu$^{1}$, Jingliang Duan$^{1}$, Yang Guan$^{1}$, Shengbo Eben Li$^{1*}$, Jianyu Chen$^{2}$% <-this % stops a space
\thanks{This study is supported by International Science \& Technology Cooperation Program of China under 2019YFE0100200, Tsinghua-Toyota Joint Research Institute Cross-discipline Program and Xilinx.}% <-this % stops a space
\thanks{$^{1}$State Key Laboratory of Automotive Safety and Energy, School of Vehicle and Mobility, Tsinghua University, Beijing, China  ${}^{*}$ Corresponding author {\tt\small lishbo@tsinghua.edu.cn}}%
\thanks{$^{2}$Institute for Interdisciplinary Information Sciences, Tsinghua University, Beijing, China {\tt\small jianyuchen@tsinghua.edu.cn}}%
}
\begin{document}
\maketitle
\thispagestyle{empty}
\pagestyle{empty}

\begin{abstract}
 Safety is essential for reinforcement learning (RL) applied in real-world tasks like autonomous driving. Chance constraints which guarantee the satisfaction of state constraints at a high probability are suitable to represent the requirements in real-world environment with uncertainty. Existing chance constrained RL methods like the penalty method and the Lagrangian method either exhibit periodic oscillations or cannot satisfy the constraints. 
 In this paper, we address these shortcomings by proposing a separated proportional-integral Lagrangian (SPIL) algorithm. Taking a control perspective, we first interpret the penalty method and the Lagrangian method as proportional feedback and integral feedback control, respectively. Then, a proportional-integral Lagrangian method is proposed to steady learning process while improving safety. To prevent integral overshooting and reduce conservatism, we introduce the integral separation technique inspired by PID control. Finally, an analytical gradient of the chance constraint is utilized for model-based policy optimization. The effectiveness of SPIL is demonstrated by a narrow car-following task. Experiments indicate that compared with previous methods, SPIL improves the performance while guaranteeing safety, with a steady learning process. 

\end{abstract}

\section{Introduction}
Reinforcement Learning (RL) has shown exceptional successes in a variety of domains, from video games \cite{vinyals2019grandmaster,Mnih2015HumanlevelCT,Hessel2018RainbowCI} to robotics \cite{kurutach2018model,haarnoja2018soft}. As a self-learning method, RL is promising to reduce the massive engineering efforts in autonomous driving.
%%JC.2.8: We might need a sentence here explaining why studying RL for AD is valuable, e.g, help handle corner cases with an data-driven approach, reduce massive engineering efforts... We are not just studying this because there is a growing interest.
In recent years, there has been a growing interest towards RL in autonomous driving community, such as adaptive cruise control \cite{lin2019longitudinal}, lane-keeping \cite{duan2020hierarchical}, trajectory tracking \cite{Mu2020MixedRL} and multi-vehicle cooperation \cite{Guan2020CentralizedCF}. However, despite achieving decent performance, these RL methods mostly lack explicit safety constraints, which significantly limits their application in safety-critical autonomous driving.

Recently, some RL researchers begin to investigate including different forms of safety constraints in RL algorithms to improve safety for real-world applications \cite{chow2017risk,achiam2017constrained,yang2019projection,garcia2015comprehensive}. 
%%JC.2.8: Change "safety guarantees" to some milder statement, e.g, improve safety, handle safety constraints... "guarantee" is too strong for now, and the cited reference actually did not provide such guarantees.
One of the most popular forms is the chance constraint, which constrains the possibility of the control policy violating the state constraint below a given level\cite{Peng2020ModelBasedAW, Paternain2019LearningSP, chow2017risk}. Chance constraint gives an intuitive and quantitative measure of the safety level of the control policy, so it is suitable to represent the safety demands in real-world systems with uncertainty.
%%JC.2.8: We need to explain a bit here why chance constraint is good so we use it, just stating it's popular is not convincing enough.

\begin{figure}[hbt]
    \centering
    \subfigure[]{
        \includegraphics[width=0.225\textwidth]{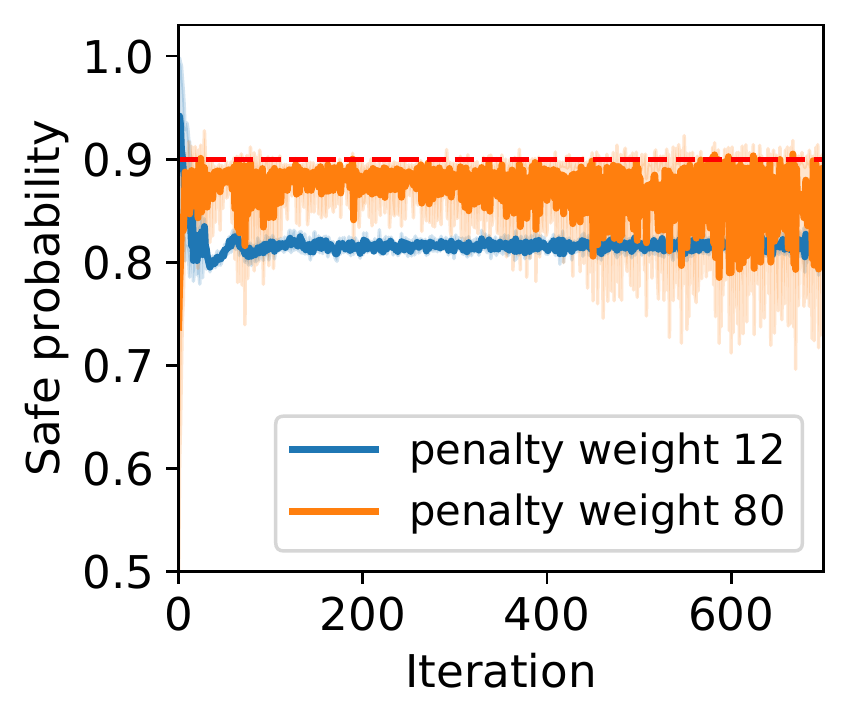}
        \label{demo_penalty0.9}
    }
    \subfigure[]{
	\includegraphics[width=0.225\textwidth]{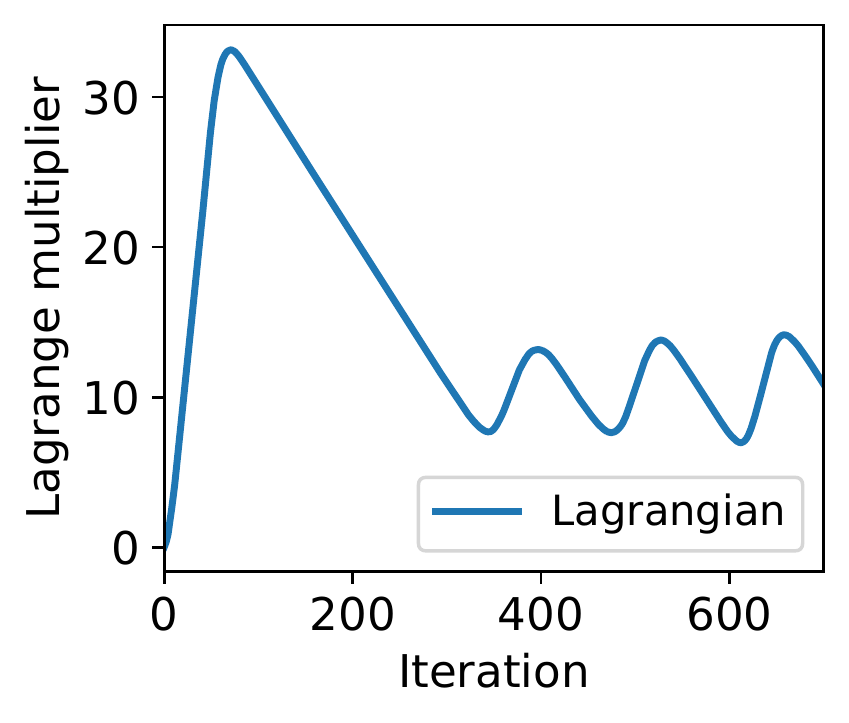}
        \label{demo_Lag0.9}
    }
    \caption{(a) Penalty method exhibits oscillations and violates the constraint. (b) Lagrangian method exhibits overshooting and oscillations of Lagrange multiplier.}
\end{figure}

Existing strategies used to solve the chance constrained RL problems can be roughly categorized into two approaches. 
%%JC.2.8: Are you sure these two approaches capture the whole categories? If not, say "can be mainly/roughly"
The first solution is the penalty method that gives a large penalty to the objective function for violation of the constraint \cite{Guan2020CentralizedCF,Peng2020ModelBasedAW}. Although this approach is very straightforward and simple to implement, it requires the penalty weight to strike a balance between safety and performance correctly. Unfortunately, it is usually difficult to select an appropriate weight. As shown in Fig. \ref{demo_penalty0.9}, a large penalty is prone to rapid oscillations and does not converge to a safe policy, while a small penalty cannot satisfy the constraint \cite{tessler2018reward}. 
%%JC.2.8: Here in this paragraph it just states the there are oscillations, but lacks the explanations about oscillations of "what" and how it harms. This is actually what the readers really care about. 
The second approach is the Lagrangian method \cite{Paternain2019LearningSP,chow2017risk}, which is widely used in constrained optimization. Actually, it can be regarded as the penalty method with an adaptive weight, which is dynamically adjusted by safety level rather than fixed. Nevertheless, the Lagrangian method suffers from overshooting of Lagrange multiplier under tight chance constraint as shown in Fig. \ref{demo_Lag0.9}, which leads to a conservative policy. Besides, it may also have periodic oscillations, resulting from the delay between the optimization of policy and adaptation of the Lagrange multiplier \cite{wah2000improving, stooke2020responsive}.
%%JC.2.8: It would be better to also speak something about the limitations of the existing chance constraint formulation, since this corresponds to the second main contribution of this work.

To overcome the drawbacks of the two methods, we take the dynamical systems view of optimization \cite{An2018APC, stooke2020responsive} and propose a separated proportional-integral Lagrangian (SPIL) method which can fulfill the safety requirements with a steady and fast learning process. From a control perspective, the safe probability is the control output and the penalty weight is the control input. Then the penalty method can be interpreted as a proportional feedback controller, while the Lagrangian method can be interpreted as an integral feedback controller. Subsequently, the proportional-integral (PI) Lagrangian method is formulated, which integrates both methods to get their merits.
%%JC.2.8: Is the PI Lagrangian method here actually same with that in Stooke2020ResponsiveSI? If yes, change "proposed" to word like "formulated".
To prevent integral-overshooting, we draw inspiration from PID control again and introduce the integral separation technique, i.e., separate the integrator out when the feedback error is large. Such a recipe solves the integral-overshooting problem that is ignored and unsolved in similar works \cite{stooke2020responsive}. In addition, we also adopt an analytical gradient of the safe probability with the theoretical basis to solve chance constraints in a model-based framework. Finally, the experiment of a narrow car-following task demonstrates SPIL succeeds in satisfying the chance constraint while achieving best cumulative reward. The contributions of this paper are as follows,
\begin{enumerate}
\item an integral-separated proportional-integral Lagrangian (SPIL) method is proposed to solve chance constrained RL problems with better performance while satisfying the constraint.
\item an analytical gradient of safe probability is adopted for model-based policy optimization with theoretic basis.
\end{enumerate}
%%JC.2.8: It'll be better to briefly introduce the vehicle case experiment and what kind of good results we have.
The rest of this paper is organized as follows. The chance constrained RL problem is formulated in Section \ref{sec:Preliminary}. The SPIL method is proposed in Section \ref{sec:chance constrained algorithm}. The effectiveness of the method is illustrated by a narrow car-following task in Section \ref{sec:Numerical Experiment}. Section \ref{sec:Conclusion} concludes this paper.

\section{Chance Constrained RL problems}
\label{sec:Preliminary}
Considering a discrete-time stochastic system, the dynamic with the chance constraint is mathematically described as:
\begin{equation}
\begin{aligned}
&x_{t+1}=f(x_{t}, u_{t}, \xi_{t}),\\
&\xi_{t}\sim p(\xi_{t}),\\
&{\rm Pr}\left\{ \bigcap_{t=1}^{N} \left[h\left(x_{t}\right)<0\right]\right\}\ge1-\delta
\end{aligned}
\end{equation}
where $t$ is the current step, $x_{t}\in{\mathcal{X}}$ is the state, $u_{t}\in{\mathcal{U}}$ is the action, $f(\cdot,\cdot,\cdot)$ is the environmental dynamic model, $\xi_{t}\in{\mathbb{R}}^{n}$ is the uncertainty following an independent and identical distribution $p(\xi_{t})$. $h(\cdot)$ is the state constraint function defining a safe state region. We do not make assumptions about the form of $f(\cdot,\cdot,\cdot)$ and $h(\cdot)$, i.e., they can be linear or nonlinear. Note that here the safety constraint takes the form of a joint chance constraint with $1-\delta$ as the required threshold. This form is extensively used in stochastic systems control \cite{Mesbah2016StochasticMP}. Intuitively, it can be interpreted as the probability of the plant staying within a safe region over the finite horizon $N$ is at least $1-\delta$. For simplicity, we only consider one constraint.

The objective of chance constrained RL problems is to maximize the expectation of cumulative reward $J$, while constraining the safe probability $p_s$:
\begin{equation}
\begin{aligned}
&\max _{\pi} J\left(\pi\right)={\mathbb{E}_{x_0,\xi}}\left\{\sum_{t=0}^{\infty} \gamma^{t} r\left(x_{t}, u_{t}\right)\right\}\\
&\text { s.t. }p_s(\pi)={\rm Pr}\left\{ \bigcap_{t=1}^{N} \left[h\left(x_{t}\right)<0\right]\right\}\ge1-\delta
\label{CCRL problem}
\end{aligned}
\end{equation}
where $r(\cdot,\cdot)$ is the reward function, $\gamma\in(0,1)$ is the discounting factor, ${\mathbb{E}_{x_0,\xi}}(\cdot)$ is the expectation w.r.t. the initial state $x_0$ and uncertainty $\xi$. $\pi$ is the control policy, i.e., a deterministic mapping from state space ${\mathcal{X}}$ to action space ${\mathcal{U}}$ with parameters $\theta$, i.e., $u_{t}=\pi(x_{t};\theta)$.

\section{Separated PI Lagrangian for Chance Constrained RL Problems}
\label{sec:chance constrained algorithm}
In this section, we will elaborate on the SPIL method for chance constrained problems. Besides, we also introduce an analytical gradient of safe probability and update the policy in a model-based mechanism. 
\subsection{Separated PI Lagrangian method}\label{SPIL}
The PI Lagrangian method comes from a control view of the penalty method and the traditional Lagrangian method. It considers the penalty method as a proportional feedback controller and the traditional Lagrangian method as an integral feedback controller, which can be integrated together and lead to a PI Lagrangian method. To see this, we first review the penalty method and traditional Lagrangian method.
The penalty method adds a quadratic penalty term in the objective function to force the satisfaction of the constraint:
\begin{align}
\label{penalty}
    &\max_\pi J(\pi) - \frac{1}{2}\alpha_p \left((1-\delta- p_s(\pi))^+\right)^2
\end{align}
where $\alpha_p>0$ is the penalty weight, $(\cdot)^+$ means $\max(\cdot,0)$.
This unconstrained problem is solved by gradient ascent
\begin{equation}
\label{penalty_update}
    \theta^{k} \leftarrow \theta^{k-1}+\alpha_{\theta}(\nabla_{\theta}J^{k}+\alpha_p (1-\delta - p_s^k)^+ \nabla_{\theta}p_s^{k})
\end{equation}
where $k$ means $k$-th iteration, $\alpha_{\theta}>0$ is the learning rate.

As for the traditional Lagrangian method, it first transforms the original chance constrained problem \eqref{CCRL problem} into an dual problem by introduction of a the Lagrange multiplier $\lambda$ \cite{boyd2004convex}:
\begin{align}
\label{maxmax}
    &\max_{\lambda\ge0}\max_\pi\mathcal{L}(\pi,\lambda)=J(\pi)-\lambda\left(1-\delta- p_s(\pi)\right)
\end{align}

The problem \eqref{maxmax} is solved by iteratively updating the Lagrange multiplier and primal variables:
\begin{equation}
\label{dual_update}
    \lambda^{k}\leftarrow (\lambda^{k-1} +\alpha_{\lambda}(1-\delta- p_s^{k}))^+
\end{equation}
\begin{equation}
\label{primal_update}
    \theta^{k} \leftarrow \theta^{k-1}+\alpha_{\theta}(\nabla_{\theta}J^{k}+\lambda^k \nabla_{\theta}p_s^{k})
\end{equation}
where $\alpha_{\lambda}>0$ is the learning rate. Comparing the policy update rule of penalty method \eqref{penalty_update} with that of Lagrangian method \eqref{primal_update}, one may find they are surprisingly similar. Both gradients is the weighted sums of  $\nabla_{\theta}J^{k}$ and $\nabla_{\theta}p_s^{k}$. The only difference lies in that the weight $(1-\delta + p_s^k)^+$ in penalty method is the one-step constraint violation, while the weight $\lambda$ in Lagrangian method is the cumulative constraint violation as \eqref{dual_update} shows. This insight builds the bridge between optimization and feedback control. One can view the optimization as dynamic systems control, where the weight of $\nabla_{\theta}p_s$ is the control input, $p_s$ is the control output and $1-\delta$ is the desired output. Consequently, the penalty method becomes a proportional controller with coefficient $\alpha_p$, while the Lagrangian method becomes an integral controller with coefficient $\alpha_\lambda$. Considering constrained optimization in such a control perspective, one can immediately understand the merits and faults of these two methods. For the penalty method, a large penalty $\alpha_p$ is prone to oscillations, while a small penalty leads to steady-state errors, i.e., not satisfying the constraint. For the Lagrangian method, it suffers from periodic oscillations from a delayed feedback. 

Subsequently, we naturally formulate a proportional-integral Lagrangian method to realize fast and steady learning process with no steady-state error. The update rule is a combination of previous two methods:
\begin{align}
\label{P_update}
    &\Delta^{k} \leftarrow 1-\delta-p_s^{k} \\
\label{I_update}
    &I^{k} \leftarrow (I^{k-1} + \Delta^{k})^+ \\
% \label{D_update}
%     &\Psi^{k} \leftarrow (p_s^{k-1} - p_s^{k})^+ \\
\label{lambda_update}
    &\lambda^{k} \leftarrow (K_P\Delta^k+K_II^k)^+ \\
\label{PI_update}
    &\theta^{k} \leftarrow \theta^{k-1}+\alpha_{\theta}(\nabla_{\theta}J^{k}+\lambda^k \nabla_{\theta}p_s^{k})
\end{align}

where $\Delta, I$ are proportional and integral values, respectively, with $K_P,K_I$ denoting their corresponding coefficients. The proportional term $\Delta$ serves as an immediate feedback of the constraint violation. The integral term $I$ eliminates the steady-state error at convergence. In such a framework, the penalty method and traditional Lagrangian method can be regarded as two special cases of PI Lagrangian with $K_P>0, K_I=0$ and $K_P=0,K_I>0$, respectively. Actually, the proportional and integral terms together will achieve better performance in RL, just as the PI controller works well in control area. 

However, if the chance constraint is very tight and the initial policy is relatively unsafe, the integral terms usually increase rapidly since $\Delta$ is large, which will cause the overshooting of $\lambda$. With a large $\lambda$ in \eqref{PI_update}, the policy tends to become extremely conservative since the weight of $\nabla_{\theta}p_s^{k}$ is relatively large. Even worse, since the maximal safe probability is 1, the overshooting and conservatism problems will not recover by themselves. For e.g., if $1-\delta=0.999, p_s=1.0$ and the $\lambda^k$ is already overshooting, the integral term $I$ only decreases very slowly with the speed of $\Delta^{k}=-0.001$. Therefore, the policy optimization in such a case will be decelerated. This challenge is also not well recognized and resolved in previous similar works like \cite{stooke2020responsive}. 
In this paper, we draw inspiration from some anti-saturation methods in PID control \cite{Jia2019AnIP}, and introduce the integral separation technique. It reshapes the integrator in \eqref{I_update} into:
\begin{equation}
\begin{aligned}
\label{SI_update}
    &I^{k} \leftarrow (I^{k-1} + K_S\Delta^{k})^+, \\
    &K_S=
    \begin{cases}
    0& \text{$\varepsilon_1<\Delta^{k}$} \\
    \beta& \text{$\varepsilon_2<\Delta^{k}<\varepsilon_1$} \\
    1& \text{$\Delta^{k}<\varepsilon_2$}
    \end{cases}
\end{aligned}
\end{equation}
where $K_S$ is the separation function, $1>\beta>0, \varepsilon_1>\varepsilon_2>0$ are the parameters. Obviously, the piecewise function $K_S$ separates the integrator out or slows it down if the error is relatively large. Such a recipe prevents the occurrence of integral-overshooting, greatly improving the performance under tight constraint as shown in our experiments.  

% First, notice when $\lambda$ is large, the update in \eqref{primal_update} can cause dramatic change in parameters $\theta$, leading to instability. To maintain a relatively consistent step size, the update rules in \eqref{primal_update} is re-scaled:

\begin{figure}[htbp]
\centering 
\centerline{\includegraphics[width=0.5\textwidth]{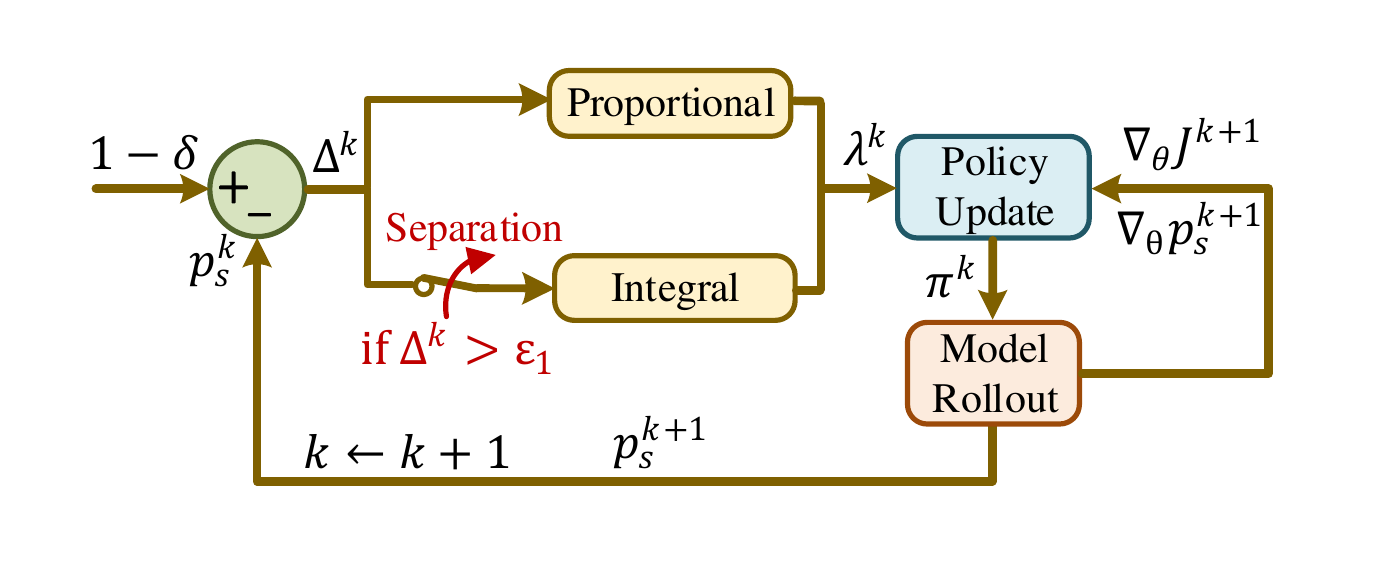}}
\caption{The framework of SPIL method.}
\label{fig:SPIL algorithm}
\end{figure}

The framework of proposed method is summarized Fig. \ref{fig:SPIL algorithm}.
% \begin{equation}
%     \theta^{k} \leftarrow \theta^{k-1}+\frac{\alpha_{\theta}}{1+\lambda^k}(\nabla_{\theta}J^{k}+\lambda^k \nabla_{\theta}p_s^{k})
% \end{equation}

\subsection{Analytical Gradient for Safe Probability}
In the previous subsection, we have derived the main update rules of our SPIL method. The following parts discuss about how to calculate $p_s$ and $\nabla_\theta p_s$ in the update equations. 

The safe probability $p_s$ in the update rules can be directly estimated through Monte-Carlo sampling. To be specific, we rollout $M$ trajectories with current policy $\pi$ through the dynamic model. Suppose there are $m$ safe trajectories, then the safety probability is $p_s\approx \frac{m}{M}$.  Note that this rollout procedure will not impose much extra computation burden since these trajectories are also necessary for the update of actor-critic as we will discuss in \ref{actor-critic}.

However, it turns out that the gradient $\nabla_\theta p_s(\pi)$ is rather difficult to compute, which is also a major challenge in chance constrained problems \cite{Mesbah2016StochasticMP, Geletu2019AnalyticAA}. Previous researchers in chance constrained RL usually replace $\nabla_\theta p_s(\pi)$ with the gradient of a lower bound of $p_s$ without sufficient theoretical guarantees \cite{Paternain2019LearningSP, Peng2020ModelBasedAW}. In this paper, we introduce an analytical approximated gradient with theoretical basis \cite{Geletu2019AnalyticAA}. To the best of our knowledge, this is the first time such a gradient is used in chance constrained RL. 

We first define an indicator-like function $\phi(x,\alpha)$:
\begin{equation}
\begin{aligned}
    &\phi(x,\tau)=\frac{1+a_1\tau}{1+a_2\tau\exp(-\frac{x}{\tau})}, \\
    &0<a_2<\frac{a_1}{1+a_1},  0<\tau<1
\end{aligned}
\end{equation}
where $\tau,a_1,a_2$ are the parameters. The expected production $\Phi(\pi,\tau)$ is defined as:
\begin{equation}
    \Phi(\pi,\tau)={\mathbb{E}_{x_0,\xi}}\left\{\prod_{t=1}^{N}\phi\left(-h(x_{t}),\tau\right)\right\}
\end{equation}
Intuitively, $\phi(x,\tau)$ can be regarded as a differentiable approximation of indicator function for constraint violation, and its expectation form $\Phi(\pi,\tau)$ approximates joint safe probability. The parameter $\tau$ controls how well the indicator function is approximated.
Under mild assumptions, the gradient of $\Phi(\pi,\tau)$ converges to the gradient of joint safe probability $p_s(\pi)$ as $\tau$ approaches $0$ \cite{Geletu2019AnalyticAA}.
\begin{equation}
\label{app_gradient}
   \lim\limits_{\tau\to0+}
\sup\limits_{\theta\in\Theta}\nabla_{\theta}\Phi(\pi,\tau)=\nabla_{\theta}p_s(\pi)
\end{equation}
where $\Theta$ is a ball in the policy parameter space. The equation \eqref{app_gradient} shows that one can use the gradient of a differentiable function $\Phi(\pi,\tau)$ to approximate $\nabla_{\theta}p_s(\pi)$ if $\tau$ is small enough. For simplicity, we do not provide more mathematical details; interested readers are recommended to refer to \cite{Geletu2019AnalyticAA} for a rigorous explanation. In practice, one only needs to pick a small fixed $\tau$ and compute $\nabla_{\theta}\Phi(\pi,\tau)$ with any autograd package, where the expectation  is substituted by sampling average. In addition, the order of magnitude of $\nabla_{\theta}J$ and $\nabla_{\theta}p_s$ are usually different. To better balance them, the gradient $\nabla_{\theta}p_s$ is re-scaled to match the scale of $\nabla_{\theta}J$:
\begin{equation}
    \nabla_{\theta}p_s \leftarrow \frac{\|\nabla_{\theta}J\|}{\|\nabla_{\theta}p_s\|}\nabla_{\theta}p_s
\end{equation}
% For simplicity, we overload the symbol $\nabla_{\theta}p_s$ to denote the re-scaled one in the rest of paper.
% It is notable that \eqref{PID_update} can be regarded as an augmented version of traditional Lagrangian \eqref{dual_update}. If we take $K_p=0$, it is exactly the same as \eqref{dual_update}.

% \begin{figure}[htbp]
% \centerline{\includegraphics[width=0.5\textwidth]{PID_Lagrangia_new.pdf}}
% \caption{PID Lagrangian methods.}
% \label{fig:exterior method}
% \end{figure}

\subsection{Model-based Actor-Critic with Parameterized Functions}
\label{actor-critic}
In this subsection, the main focus is on how to learn a parameterized policy and state-action value function in the model-based framework, where the gradient of the dynamic model will be utilized to attain an accurate ascent direction and thus improve the convergence rate compared with model-free RL algorithms \cite{deisenroth2011pilco,Shengbo2019}.

For an agent behaving according to policy $\pi$, the values of the state-action pair $(x,u)$ are defined as follows:
\begin{align}
    &Q^{\pi}(x,u)={\mathbb{E}_{\xi}}\left\{\sum_{t=0}^{\infty} \gamma^{t} r\left(x_{t}, u_{t}\right)\Big|x_0=x,u_0=u\right\}
\end{align}
Consequently, the expected cumulative reward $J$ can be expressed as a $N$-step form:
\begin{align}
\label{n-step}
    &J(\pi)={\mathbb{E}_{x_0,\xi}}\left\{\sum_{t=0}^{N-1}\gamma^{t} r\left(x_{t}, u_{t}\right)+\gamma^{N}Q^{\pi}(x_N,u_N)\right\}
\end{align}

For large and continuous state spaces, both value function and policy are parameterized, as shown in  \eqref{para}. The parameterized state-action value function with parameter $w$ is usually named the ``critic'', and the parameterized policy with parameter $\theta$ is named the ``actor'' \cite{Shengbo2019}. 
 \begin{equation}
Q(x,u) \cong Q(x,u ; w), \quad u \cong \pi(x ;\theta)
     \label{para}
\end{equation}

The parameterized critic is trained by minimizing the average square error  \eqref{td error}: 
\begin{equation}
\begin{aligned}
  J_{Q}^k={\mathbb{E}}_{x_0,\xi}\left\{\frac{1}{2}\left(Q_{\text {target}}-Q(x_0,u_0;w^k)\right)^{2}\right\}
     \end{aligned}
\label{td error}
\end{equation}
where $Q_{\text {target}} = \sum_{t=0}^{N-1} \gamma^{t} r\left(x_{t}, u_{t}\right)+\gamma^{N} Q\left(x_{N},u_{N};w^k\right)$ is the $N$-step target. Note that the rollout length $N$ is identical to the horizon of chance constraint.

The semi-gradient of the critic is
 \begin{equation}
\begin{aligned}
  \nabla_{\omega}J_Q^k&={\mathbb{E}}_{x_0,\xi}\left\{\left(Q(x_0,u_0;w^k)-Q_{\text {target}}\right) \frac{\partial Q(x_0,u_0;w^k)}{\partial w}\right\}
\end{aligned}
\label{semi-gradient of the critic}
\end{equation}

As discussed in \eqref{maxmax}, the parameterized actor aims to maximize Lagrangian function $\mathcal{L}$ via gradient ascent.  The analytical gradient $\nabla_{\theta}\mathcal{L}$ is composed of $\nabla_{\theta}J$ and $\nabla_{\theta}p_s$, which are computed via backpropagation though time with the dynamic model \cite{Shengbo2019}. In practice, they can be easily obtained by any autograd package. Finally, the pseudo-code of proposed algorithm is summarized in Algorithm \ref{alg:SPIL}. Note that, to maintain a relatively consistent step size, the update rules for $\theta$ in \eqref{PI_update} is re-scaled by $\frac{1}{1+\lambda^k}$.

\begin{algorithm}[!htb]
\caption{SPIL algorithm}
\label{alg:SPIL}
\begin{algorithmic}
\STATE Initialize $x_{0}\in \mathcal{X}$, $k=0$
\REPEAT
\STATE{Rollout $M$ trajectories by $N$ steps via dynamic model}
\STATE{Estimate safe probability}
\STATE{\quad $p_s^k\approx\frac{m}{M}$}
% \STATE{\quad $J^k=\frac{1}{M}{\sum^{M}}\left\{\sum_{t=0}^{N-1}\gamma^{t} r\left(x_{t}, u_{t}\right)+\gamma^{N}Q^{\pi}(x_N,u_N)\right\}$}
\STATE{Update $\lambda$ via PI Lagrangian rules}
\STATE{\quad  $\Delta^{k} \leftarrow 1-\delta-p_s^{k}$}
\STATE{\quad  $I^{k} \leftarrow (I^{k-1} + K_S\Delta^{k})^+$}
% \STATE{\quad  $\Psi^{k} \leftarrow (p_s^{k-1} - p_s^{k})^+$}
\STATE{\quad  $\lambda^{k} \leftarrow (K_P\Delta^k+K_II^k)^+ $}
\STATE {Update critic according to \eqref{semi-gradient of the critic}:}
\STATE{\quad  $\omega^{k} \leftarrow \omega^{k-1} + \alpha_{\omega}\nabla_{\omega}J_{Q}^k$ }
\STATE {Update actor:}
\STATE{\quad  $\theta^{k} \leftarrow \theta^{k-1}+\alpha_{\theta}\nabla_{\theta}\mathcal{L}$ }
\STATE{\quad  $\nabla_{\theta}\mathcal{L}= \frac{1}{1+\lambda^k}\left(\nabla_{\theta}J^{k}+\lambda^{k} \nabla_{\theta}p_s^{k}\right)$ }
\STATE $k\leftarrow k+1$
\UNTIL $|Q^{k}-Q^{k-1}|\le \zeta$ and $|\pi^{k}-\pi^{k-1}| \le \zeta$
\end{algorithmic}
\end{algorithm}

\section{Numerical Experiment}
%%JC.2.8: Better to split this part into subsections for clear, e.g, experiment setup, implementation details, evaluation results...
\label{sec:Numerical Experiment}
\subsection{Experiment Setup}
\begin{figure}[htbp]
\centerline{\includegraphics[width=0.4\textwidth]{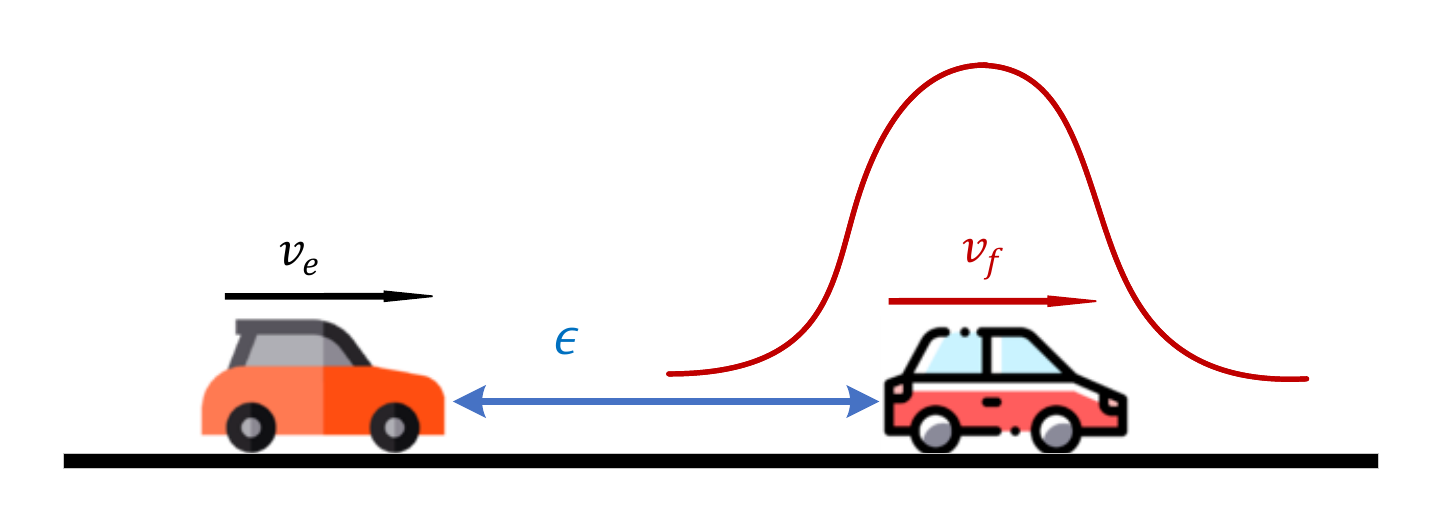}}
\caption{Car-following scenario.}
\label{fig:car-follwoing}
\end{figure}

In this section, the proposed SPIL is applied to a narrow car-following scenario as shown in Fig. \ref{fig:car-follwoing}, 
%%JC.2.8: We need to show the task is important for autonomous driving, maybe say something about how his problem is important and add related citations. Also, "aggressive" feels like a negative word, try change to some other words.
where the ego car expects to drive fast and closely with the front car to reduce wind drag \cite{gao2016robust}, while keeping a minimum gap between the two cars at a high probability. Concretely, the ego car and front car follow the kinematics model, where the front car is assumed to drive with a randomly varying velocity (e.g., due to the varying road grade, wind drag). 

The discrete-time stochastic system is  
\begin{equation}
\begin{aligned}
&x_{t+1}=A x_t+B u_t+D \xi_t \\
&A=\left[{\small{\begin{array}{ccc}
1 & 0 & 0 \\
0 & 1 & 0 \\
-T & T & 1
\end{array}}}\right],\\
&B=[T,0,0]^{\top}, \quad D=[0,T,0]^{\top}
\end{aligned}
\end{equation}

\begin{figure*}[hbt]
    \centering
    \subfigure[Cumulative reward under $90.0\%$ threshold]{
        \includegraphics[width=0.38\textwidth]{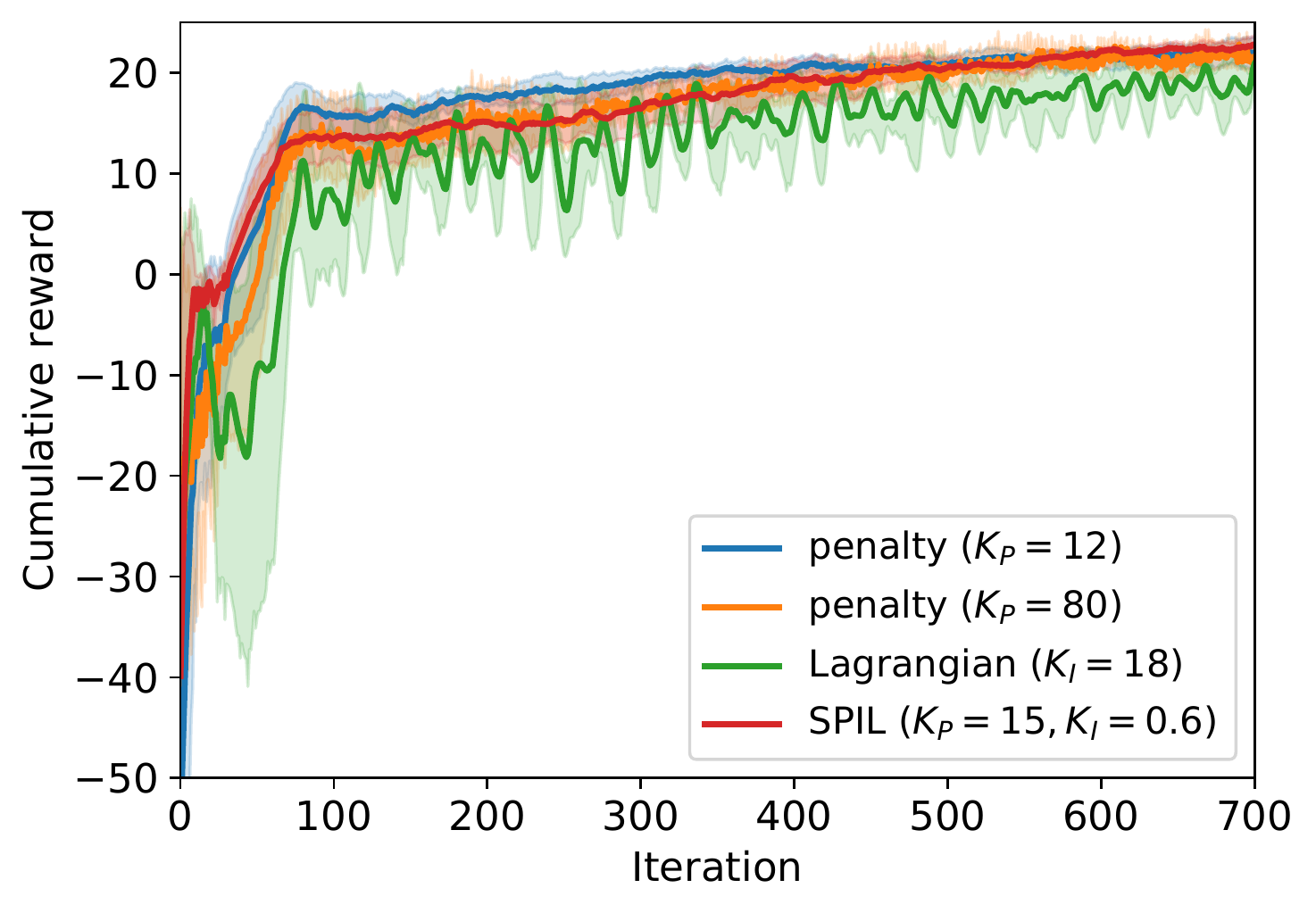}
        \label{return0.9}
    }
    \subfigure[Cumulative reward under $99.9\%$ threshold]{
	\includegraphics[width=0.38\textwidth]{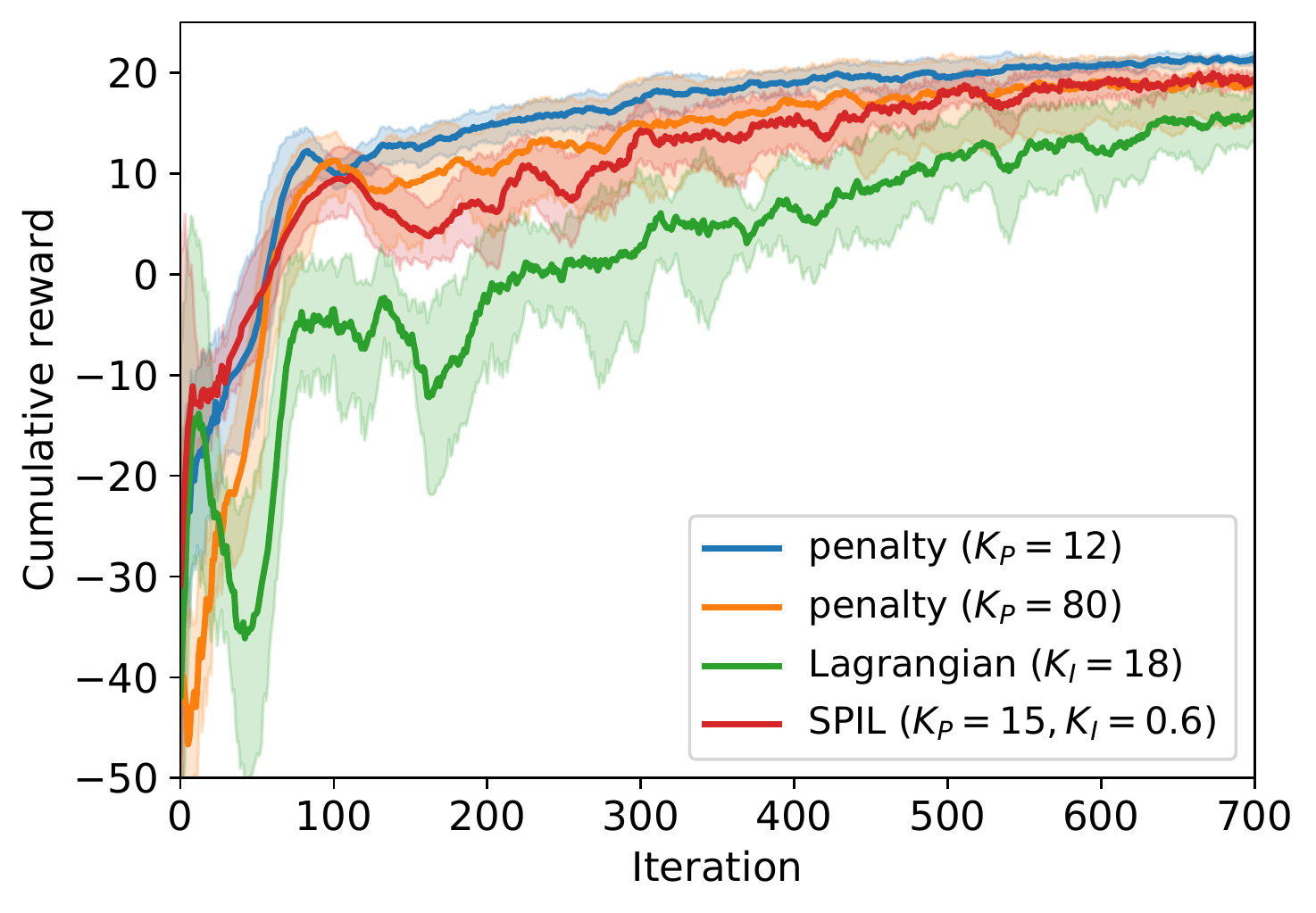}
        \label{return0.999}
    }
    \\   
    \subfigure[Safe probability under $90.0\%$ threshold]{
    	\includegraphics[width=0.38\textwidth]{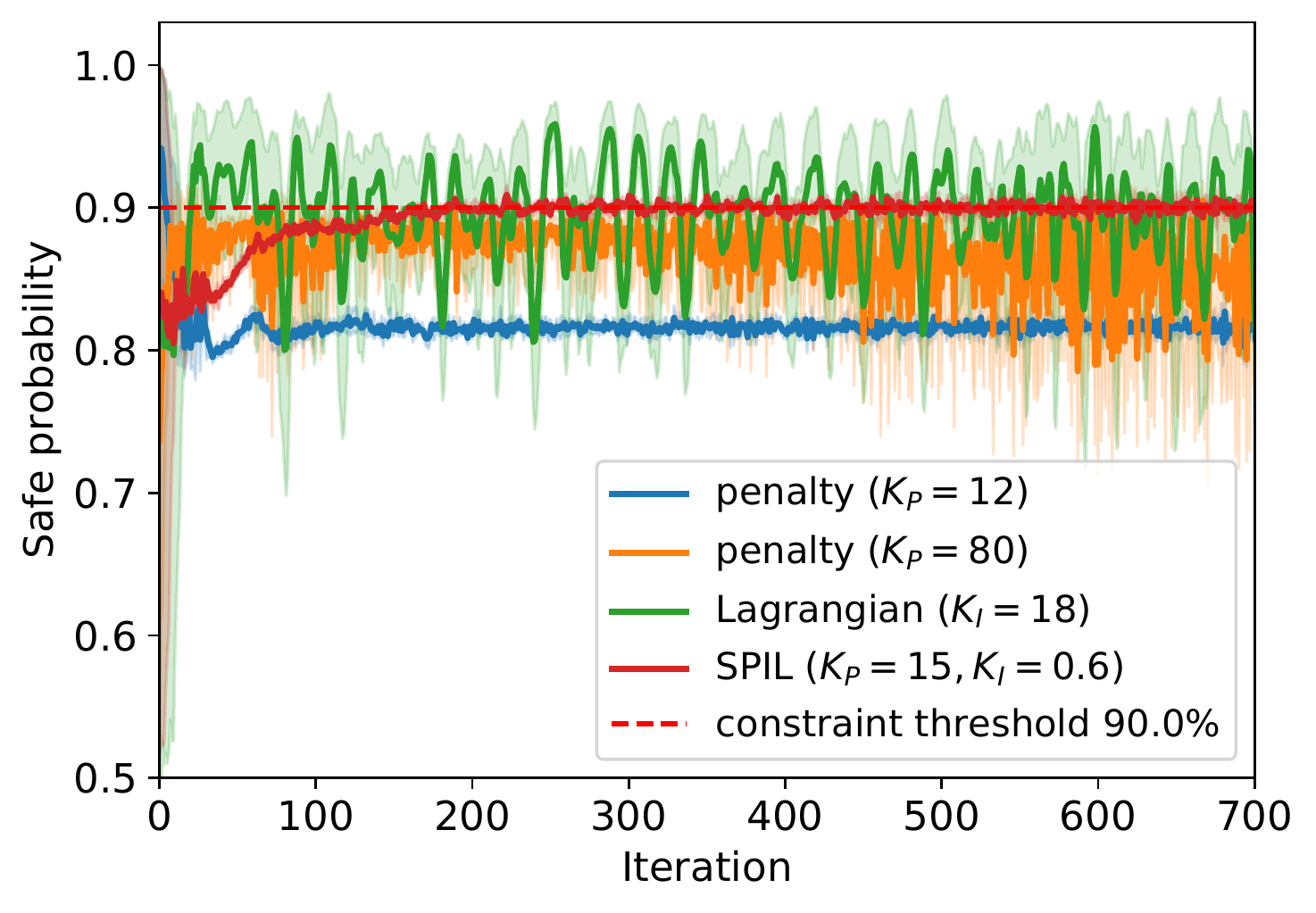}
        \label{safe0.9}
    }
    \subfigure[Safe probability under $99.9\%$ threshold]{
	\includegraphics[width=0.38\textwidth]{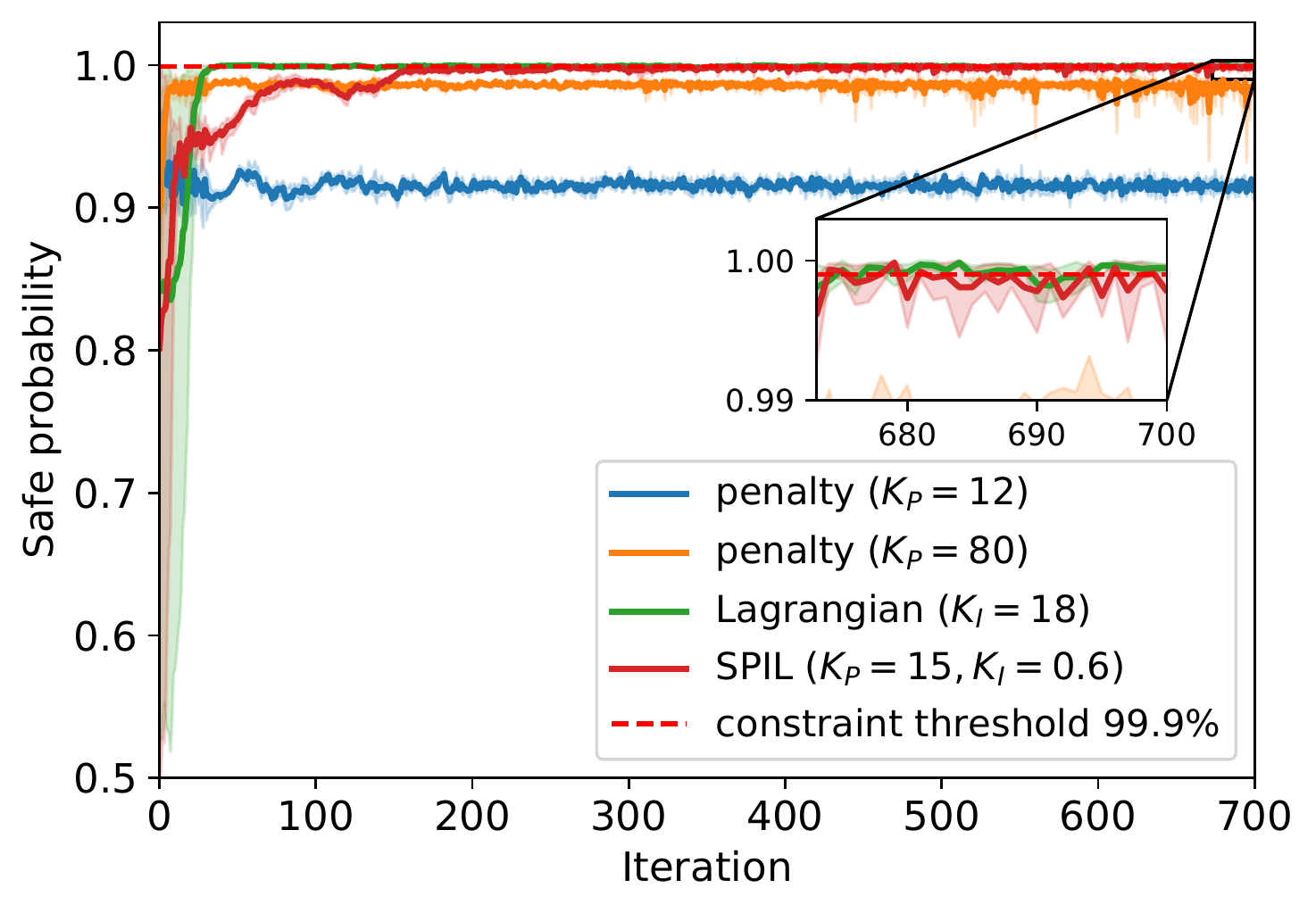}
        \label{safe0.999}
    }
%%JC.2.8: Better to add a zoom in at the areas around the 0.999 threshold, readers will be curious about the difference between SPIL and Lagrangian.
    \caption{Comparison of performance among SPIL (separated PI Lagrangian), penalty method and traditional Lagrangian method.}
    \label{performance}
\end{figure*}

The system state vector is $x =[v_e \quad v_f \quad \epsilon]^{\top}$ ,
where $v_e$ denotes the velocity of ego car, $v_f$ is the velocity of front car,
and $\epsilon$ is the gap between the two cars. The control input $u \in (-4,3)$ is the acceleration of ego car. The disturbance $\xi_t \sim \mathcal{N}(0,0.7)$ and $T=0.1 s$ is the simulation time step. With a chance constraint on the gap, the chance constrained RL problem is defined as
\begin{equation}\begin{aligned}
&\max _{\pi} \sum_{t=0}^{\infty} \gamma^{t}(0.2v_{e,t}-0.1\epsilon_t-0.02u^2_t) \\
&\text { s.t. } {\rm Pr}\left\{ \bigcap_{t=1}^{N} \left(\epsilon_{t}>2\right)\right\}\ge1-\delta
\end{aligned}\end{equation}
where $v_{e,t}$ denotes the ego car velocity at step $t$.
\subsection{Implementation Details}
We implement SPIL algorithm on the problem above. Our parameterized actor and critic are both fully-connected neural networks.  Each network has two hidden layers using rectified linear unit (ReLU) as activation functions, with 64 units per layer. We adopt the Adam method to update the networks \cite{lecun2015deep}. The main hyper-parameters are listed in Table \ref{tab:hyper}.

To demonstrate the advantages of SPIL, we compare the performance of SPIL with the penalty method (amounts to proportional-only SPIL) and traditional Lagrangian method (amounts to integral-only SPIL). The coefficients of SPIL are $K_P=15, K_I=0.6$. The penalty method is trained in two different weights $K_P=12$ and $80$. The traditional Lagrangian is trained on $K_I=18$ (we had tested on small $K_I=0.6$ but got terrible results). The cumulative reward and safe probability in horizon $N$ are compared under two chance constraint thresholds 90.0\% and 99.9\%, i.e., $\delta=0.1$ and $\delta=0.001$.

\begin{table}[hbt]
\caption{Hyper-parameters}
\begin{center}
\label{tab:hyper}
\begin{tabular}{lcc}
\hline
Parameters                       & Symbol                & Value    \\ \hline
trajectories number             & $M$                   & 4096      \\
constraint horizon              & $N$                   & 40        \\
discounting factor              & $\gamma $             & 0.99      \\
learning rate of policy network & $\alpha_{\theta}$     & 3e-4     \\
learning rate of value network  & $\alpha_{\omega}$     & 2e-4       \\ 
parameters of $K_S$             & $\beta$               & 0.3     \\
parameters of $K_S$             & $\varepsilon_1$       & 0.2      \\
parameters of $K_S$             & $\varepsilon_2$       & 0.05      \\
parameters of $\phi(\cdot)$     & $\tau$                & 1e-3      \\
parameters of $\phi(\cdot)$     & $a_1$                  & 0.45      \\
parameters of $\phi(\cdot)$     & $a_2$                  & 1      \\

\hline
\end{tabular}
\end{center}
\end{table}

\subsection{Evaluation Results}
 The learning curves are plotted in Fig. \ref{performance}, where each curve is averaged over five independent experiments. The SPIL not only succeeds to satisfy the chance constraint without periodic oscillations, but also achieves best cumulative reward among methods which meet the safety threshold. Observing the safe probability curves in Fig. \ref{safe0.9} and Fig. \ref{safe0.999}, the proposed SPIL satisfies the chance constraint in both settings. On the contrary, the penalty method with $K_P=12$ fails to achieve the required threshold due to small penalty weight. Although one can improve the penalty size and raise $K_P$ to reduce this error (i.e., set $K_P=80$), large $K_P$ also brings about rapid oscillations as a side effect, especially when the threshold is $90.0\%$. This is because with a large $K_P$, a small change of $\Delta$ will cause a dramatic change of $\lambda$ . The Lagrangian method does not have steady-state errors, but suffers from periodic oscillations under $90.0\%$ threshold. In a word, the proposed SPIL combines the advantages of integral and proportional methods, leading to a stable learning process with no steady-state errors. Interestingly, these phenomena are quite similar to conclusions in PID control, which exhibits the beauty of understanding optimization from the control perspective.
 
As for the cumulative reward shown in Fig. \ref{return0.9} and Fig. \ref{return0.999}, excluding the unsafe penalty method $(K_P=12)$, SPIL achieves the best cumulative reward in both thresholds among the other three methods, which confirms the excellent performance of SPIL. 

\begin{figure*}[hbt]
    \centering
    \subfigure[Safe probability under $99.9\%$ threshold]{
        \includegraphics[width=0.3\textwidth]{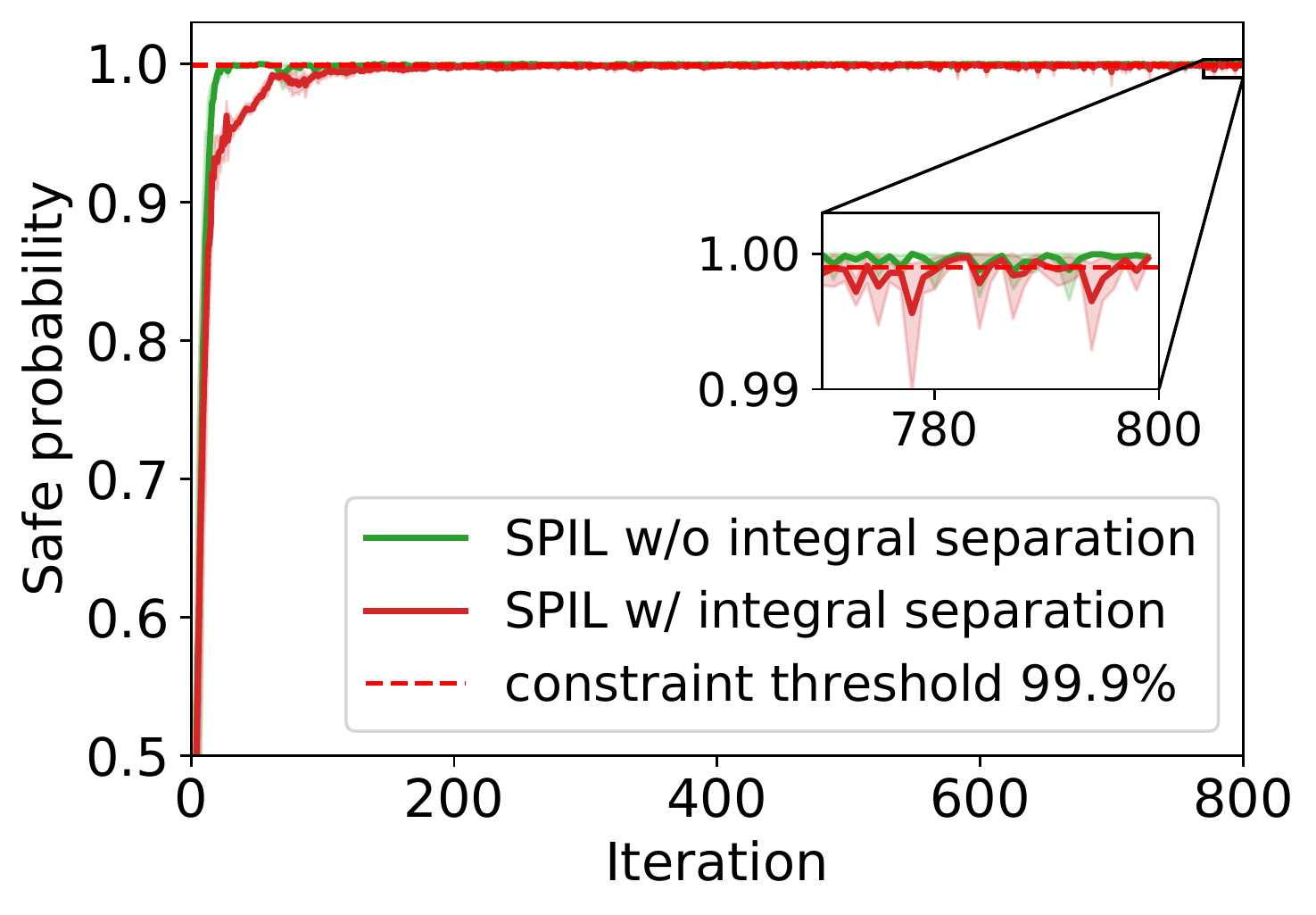}
        \label{unsafe_safety0.999}
    }
%%JC.2.8: Similarly, add a zoom in
    \subfigure[Cumulative reward under $99.9\%$ threshold]{
	\includegraphics[width=0.3\textwidth]{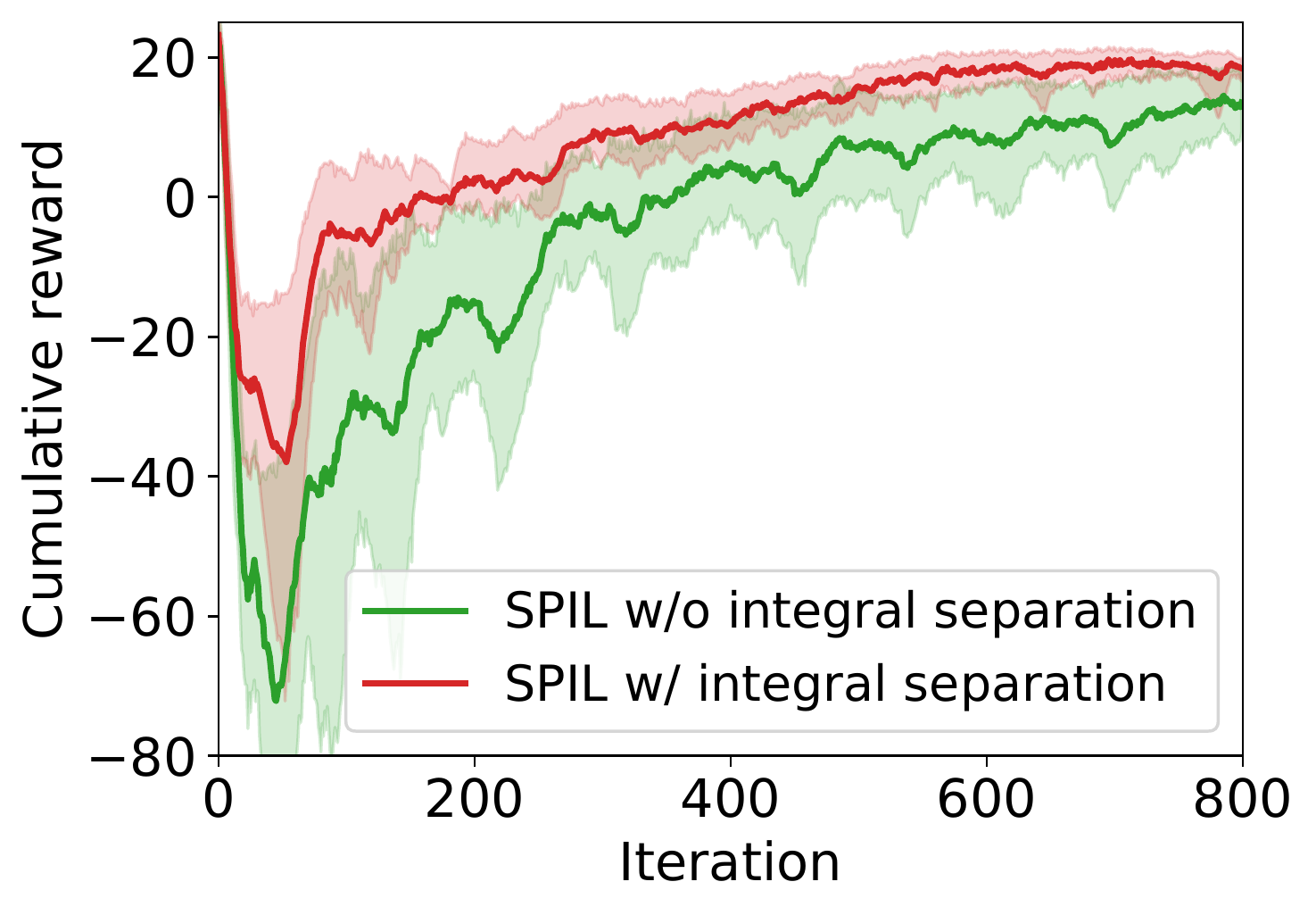}
        \label{unsafe_return0.999}
    }
    \subfigure[Integral value under $99.9\%$ threshold]{
    	\includegraphics[width=0.3\textwidth]{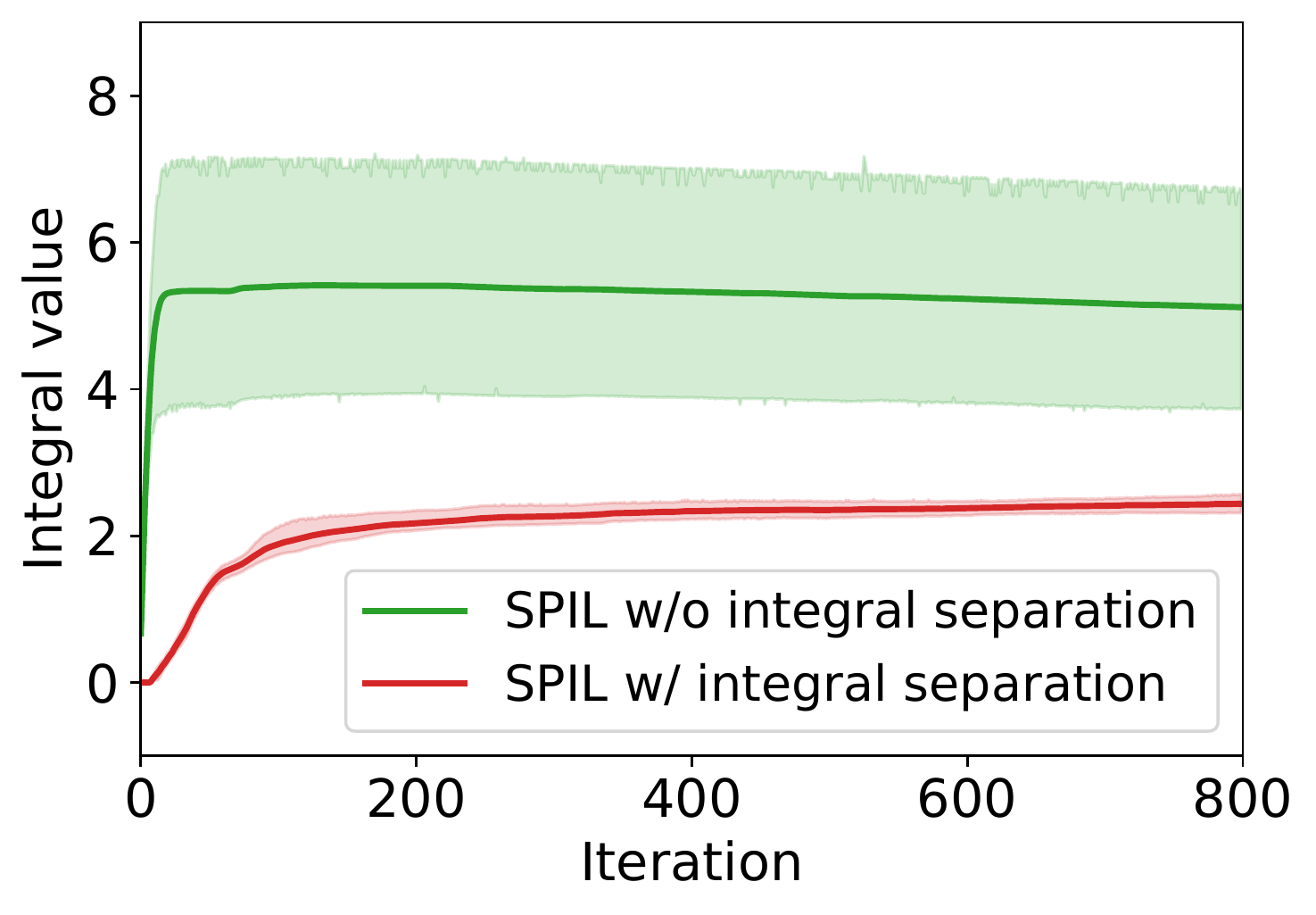}
        \label{unsafe_integral0.999}
    }
    \caption{Comparison of performance of SPIL with and without integral separation.}
    \label{unsafe_performance}
\end{figure*}

Subsequently, we demonstrate that the integral separation technique in SPIL helps to prevent integral overshooting and reduce policy conservatism. We manually select five initially unsafe random seeds, i.e., the safe probability of initial policy $p_s<0.5$, and train the policy under $99.9\%$ threshold using SPIL with and without integral separation. The learning curves of cumulative reward $J$, safe probability $p_s$, integral value $I$ are plotted in Fig. \ref{unsafe_performance}. If the integral separation is removed, the integral value $I$ in Fig. \ref{unsafe_integral0.999} will have a sharp rise at the beginning. Then the policy rapidly learns to satisfy the constraint with safe probability becoming 1. However, since $\Delta=-0.001$ in \eqref{I_update}, the decrease of $I$ is quite slow. With the excessively large $I$ and $\lambda$,  the policy keeps conservative for a long time and wins few rewards. On the contrary, with the help of integral separation, $I$ will not overshoot at the start and the policy successfully strikes a good balance between performance and safety, i.e., achieves more rewards while satisfying the constraint. Note that the results in Fig. \ref{performance} and Fig. \ref{unsafe_performance} are not comparable since the latter are conducted under manually chosen bad initial policies. 
%%JC.2.8: It is not clear at the first glimpse what conclusions we can draw from the last two paragraphs. Try to summary each conclusion with one short and clear sentence and add it as a \textbf at the beginning of the paragraphs.

\section{Conclusion}
\label{sec:Conclusion}
This paper proposed the separated PI Lagrangian method for chance constrained problems.
Viewing optimization from a control perspective, SPIL adopted PI feedback to adjust the Lagrange multiplier and achieved good performance with a steady and fast learning process. Besides, integral separation was also included to prevent overshooting and reduce conservatism. Finally, we utilized an analytical gradient of safe probability for model-based policy optimization.
The benefits of SPIL were demonstrated in simulations of a narrow car-following task. It achieved more cumulative reward while satisfying the chance constraint. 
The application of SPIL to more general environmental dynamics will be investigated in the future.
%%JC.2.8: need to use past tense for the conclusion section

\bibliographystyle{ieeetr}

\end{document}